\documentclass[final,3p,times]{elsarticle}
\usepackage{float}
\usepackage{amssymb}
\usepackage{amsmath}
\usepackage{url}

\usepackage{booktabs}
\usepackage{multirow} 
\usepackage{array} 
\usepackage{siunitx} 
\journal{Computer Methods in Applied Mechanics and Engineering}

\begin{document}

\begin{frontmatter}



\title{General Explicit Network (GEN): A novel deep learning architecture for solving partial differential equations} 


\author[1,2]{Genwei Ma}
\ead{magenwei@126.com}
\author[3]{Ting Luo}
\author[4]{Ping Yang}
\author[5]{Xing Zhao\corref{cor1}}
\ead{zhaoxing\_1999@126.com}
\address[1]{National Center for Applied Mathematics Beijing, Capital Normal University, No. 105 West Third Ring Road North, Beijing, 100048, China}
\address[2]{The Academy for Multidisciplinary Studies, Capital Normal University, No. 105 West Third Ring Road North, Beijing, 100048, China}
\address[3]{Academy of Information Network Security, People’s Public Security University of China, No.1 Muxidi Nanli, Beijing, 100038, China}
\address[4]{Institute of Nuclear and New Energy Technology, Tsinghua University, Haidian  District, Beijing, 100084, China}
\address[5]{School of Mathematical Sciences, Capital Normal University, No. 105 West Third Ring Road North, Beijing, 100048, China}
\cortext[cor1]{Corresponding author}


%
\begin{abstract}
Machine learning, especially physics-informed neural networks (PINNs) and their neural network variants, has been widely used to solve problems involving partial differential equations (PDEs). The successful deployment of such methods beyond academic research remains limited. For example, PINN methods primarily consider discrete point-to-point fitting and fail to account for the potential properties of real solutions. The adoption of continuous activation functions in these approaches leads to local characteristics that align with the equation solutions while resulting in poor extensibility and robustness. A general explicit network (GEN) that implements point-to-function PDE solving is proposed in this paper. The "function" component can be constructed based on our prior knowledge of the original PDEs through corresponding basis functions for fitting. The experimental results demonstrate that this approach enables solutions with high robustness and strong extensibility to be obtained.
\end{abstract}
\begin{keyword}
partial differential equations, physics-informed neural networks
\end{keyword}

\end{frontmatter}



\section{Introduction}\label{sec1}
In the fields of science and engineering, partial differential equations (PDEs) are widely employed in their simplest mathematical forms to describe the behaviours of complex systems spanning fluid dynamics, electromagnetism, and quantum mechanics \cite{wang2024pinn}. These equations establish a foundational framework for understanding and simulating natural phenomena. However, the traditional numerical methods for solving PDEs, which rely on progressive computations that extrapolate from the initial conditions in a step-by-step manner, often demand substantial computational resources and time. Moreover, although these traditional methods achieve high accuracy and are supported by rigorous error and stability analyses, their computational costs scale exponentially with the dimensionality of the underlying PDEs, resulting in the curse of dimensionality.

In recent years, machine learning models, especially deep neural networks (DNNs), have emerged as transformative tools, demonstrating remarkable advancements in both computational speed and efficiency for solving PDEs, thereby unlocking new perspectives and possibilities for scientists and engineers \cite{karniadakis2021physics, thuerey2021physics, cuomo2022scientific, vinuesa2022enhancing, brunton2024promising}. In fact, the theoretical foundation for solving PDEs using DNNs stems from the universal approximation theorem \cite{hornik1989multilayer}, which asserts that DNNs can theoretically approximate any continuous function. The contemporary machine learning approaches for solving PDEs predominantly fall into two methodological categories: neural operator learning frameworks \cite{kovachki2023neural,anandkumar2020neural,rosofsky2023applications} and physics-informed neural network (PINN) variants enhanced with physical constraints. The former paradigm focuses on learning differential operators through architectures such as deep operator network (DeepONet) \cite{lu2021learning,li2023phase,li2024tutorials}, the low-rank neural operator (LNO), the multipole graph neural operator (MGNO) \cite{li2020multipole}, the Fourier neural operator (FNO) \cite{lifourier}, and the Laplace neural operator \cite{cao2024laplace}, enabling efficient solutions to be obtained for parametric PDEs with shared mathematical structures but varying coefficients. While these operator-based learning methods demonstrate remarkable generalization capabilities once trained, their development processes pose three primary challenges: 1) their heavy reliance on extensive numerical simulation datasets for training, 2) the inherent neglect of the governing physical principles that are encoded in PDEs, and 3) the substantial computational overhead derived from both data acquisition and network optimization steps. These limitations have become particularly apparent since the emergence of physics-informed learning paradigms, leading to diminished interest in purely data-driven operator-based learning approaches.

The latter paradigm originates from the ground-breaking PINN framework introduced by Raissi et al. (2019) \cite{raissi2019physics}, which established a novel computational paradigm by embedding PDE formulations directly into neural network architectures. This methodology synergistically integrates supervised learning with physical constraints through autodifferentiation mechanisms of deep learning, enabling not only efficient numerical solutions but also physically consistent solutions to be obtained for PDEs. The framework has catalysed transformative advancements in computational mathematics and engineering physics \cite{cai2021physics,wu2023comprehensive,sharma2023review}, garnering significant interdisciplinary recognition, as evidenced by its 20017 citations (Google Scholar, 16 February 2026).

However, despite the demonstrated precision and computational efficiency of the contemporary deep learning techniques in terms of solving PDEs, emerging scepticism within the scientific community has raised methodological concerns about machine learning approaches. A seminal meta-analysis conducted by McGreivy and Hakim \cite{mcgreivy2024weak} involved systematically evaluating 82 studies on machine learning-based PDE solvers in fluid dynamics scenarios, revealing critical flaws in the current benchmarking practices. Furthermore, PINN methods generally fail to converge to reasonable approximations \cite{chuang2023predictive}, even for simple toy problems \cite{wang2022and,krishnapriyan2021characterizing,basir2022critical}; thus, they do not appear to be superior to alternative approaches such as discrete grid-based methods \cite{chuang2022experience, karnakov2024solving}.
Furthermore, Brandstetter \cite{brandstetter2025envisioning} fundamentally questioned whether machine learning offers substantive advantages beyond selective speed improvements, emphasizing the need for comprehensive evaluation frameworks that assess accuracy, generalizability, and computational costs across diverse physical regimes. Indeed, the application of machine learning to PDE solvers remains a solution looking for a problem \cite{M2025Machine}.

The current research addressing the limitations of PINNs focuses primarily on optimizing ill-posedness in domain-specific applications. Chen et al. \cite{chen2025pf} proposed PF-PINNs, which employ normalization techniques to mitigate spatiotemporal scale discrepancies, coupled with a neural tangent kernel (NTK)-based \cite{jacot2018neural,wight2020solving,huang2021fl} adaptive weighting strategy to balance multitask loss terms for solving coupled Allen-Cahn and Cahn-Hilliard phase field equations. The Sinc Kolmogorov-Arnold network (SincKAN) \cite{yu2024sinc,liu2024kan} architecture replaces conventional activation functions with Sinc functions, improving upon the high-frequency feature detection capabilities of the former; this approach has been successfully applied to phonon Boltzmann equations. In the context of Fourier neural networks, researchers have observed a spectral bias where networks exhibit faster convergence towards low-frequency solution components than they do toward high-frequency components, with no guaranteed convergence to the high-frequency modes even after extensive iterations. The FNO \cite{li2020fourier,qi2024gabor} and Fouier PINNs \cite{sallam2023use,wang2021eigenvector,jin2024fourier, bounnah2025physics,song2023simulating} have demonstrated effectiveness in fluid dynamics applications, including two-phase (subsurface oil/water) flow PDEs \cite{zhang2022fourier} and seismic wave equations involving variable velocity models \cite{song2023simulating,wei2022small}. Physics-informed neural operators (PINOs) integrate FNOs with physical constraints to achieve high-precision solution operator approximations under zero-shot superresolution conditions, achieving 10-fold acceleration over conventional PINNs.

Although various neural architectures have provided promising experimental results in different domains, their holistic structures and intrinsic properties cannot be fundamentally optimized through network training. To address this limitation at its root, comprehensive modifications to the network architecture and design paradigm are needed. In reality, DNN-based approaches, including PINNs, essentially employ neural networks to learn a potential closed-form solution that remains valid only within specific domain intervals. Two critical aspects warrant clarification.
\begin{itemize}
\item Closed-form solution: For a DNN with inputs of $x=(x_1,x2,\cdots, x_n)^T, t$ and an output $u$, the corresponding mathematical representation can be expressed as follows:
\[u = \phi(W^L\delta(W^{L-1}\cdots \delta(w^1(x,t)+b^1)+b^{L-1})+b^L)\]
where $\phi$ and $\delta$ denote activation functions. For PDE-related problems, differentiable functions such as $tanh(\cdot)$ are typically employed. Thus, the network essentially constructs a differentiable equation that approximates the PDE solution.

\item Domain validity: We must examine how such single-function representations capture complex functional behaviours within specified intervals. Our analysis suggests that conventional DNNs require substantial parameters to accommodate pointwise inputs during training, resulting in weak neighbourhood correlations between adjacent points. Consequently, while achieving pointwise approximation accuracy within their training domains, these models often exhibit catastrophic fitting failures in extrapolation regions beyond the coverage area of the training data. We characterize this fitting paradigm as a point-to-point approximation scheme.
\end{itemize}

We emphasize that performing pointwise fitting during training, without enforcing explicit constraints to maintain functional continuity or topological consistency between adjacent regions, results in weak interneighbourhood correlations in DNNs. This ultimately leads to solutions with poor extensibility, low robustness, and compromised stability in the learned functional representations.

To transcend the aforementioned weak interdomain correlations, we re-examine PDE solution representations through dual mathematical paradigms, i.e., closed-form analytical expressions and series-expansion representations, with each paradigm possessing distinct advantages and limitations. Closed-form solutions ($u = \mathcal{H}(x,t)$) offer intuitive mathematical elegance that explicitly reveals physical properties (e.g., stability and extensibility), facilitating theoretical analyses and rapid computational implementations. However, their applicability remains constrained to specific equation types with simple boundary conditions, often failing to provide explicit solutions for complex systems. Conversely, series-expansion methods ($u=\sum a_{ij}f_i(x)g_j(t)$), e.g., Fourier series and power series, demonstrate universal adaptability through basis function decomposition, effectively approximating nonlinear and variable-coefficient problems for which analytical solutions have proven elusive. While truncation errors and diminished physical interpretability persist as limitations, each basis function inherently encodes global structural information through its spectral characteristics.

\begin{figure}
\centering
\includegraphics[width=5in,height=1.8in]{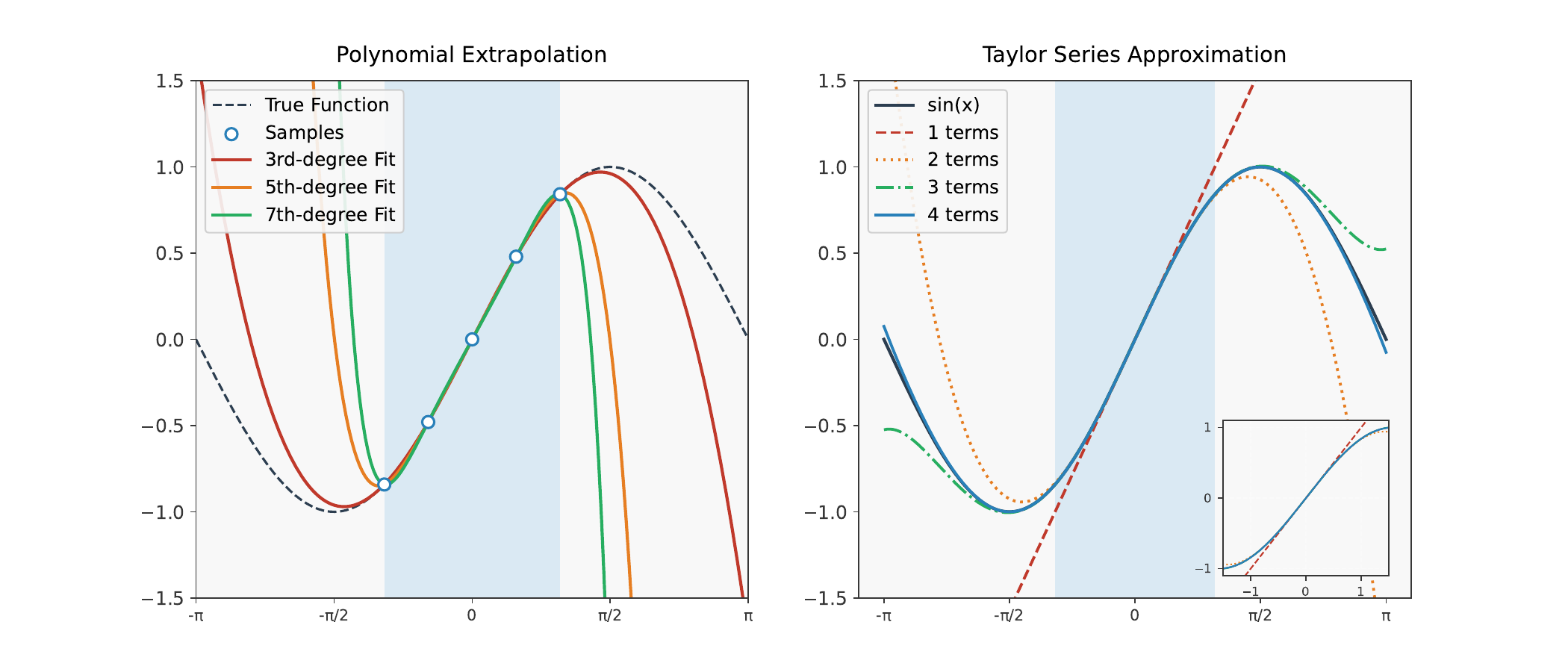}
\put(-330,120){\bf{\Large{a}}}
\put(-180,120){\bf{\Large{b}}}
\caption{This comparative study demonstrates two approaches for approximating $y=sin(x)$. (a): Explicit polynomial fitting reveals a critical trade-off: low-degree polynomials (e.g., 3rd-degree, red) fail to capture the curvature of the function within the sampled region $(x \in [-1, 1])$, whereas higher-degree fitting (5th-7th degrees, orange/green) provides improved in-sample accuracy at the cost of severe overfitting and unstable extrapolation (wild oscillations beyond $|x| > 1$). (b): Taylor series expansion exhibits progressive refinement: lower-order approximations (1-2 terms, red/brown) approximate the core trend, whereas higher-order expansions (3-4 terms, green/blue) systematically enhance both the local accuracy and global extrapolation stability of the system by aligning derivative constraints with the intrinsic physics of $sin(x)$. These divergent extrapolation behaviours underscore the superiority of series expansions over purely empirical polynomial regression schemes.}
\end{figure}

In contrast with the conventional DNN approaches that seek monolithic function approximations, we observe fundamental incompatibility with analytical solution properties such as stability preservation, which explicitly reveal physical properties. This motivates our investigation into neural architectures that mimic series-expansion mechanisms. Therefore, we propose a more general and explicit point-to-function network architecture, which we call a generalized explicit network (GEN). Formally analogous to series-expansion representations, the GEN synthesizes the final solution through a prescribed combination of basis functions. That is,
\[u=\mathcal{K}((f_1(x),\cdots, f_m(x),(g_1(t),\cdots,g_n(t)))\]
where the following hold.
\begin{itemize}
\item The operator $\mathcal{K}$ represents a composition operator for basis functions, which is analogous to the linear superposition mechanism in series expansions. To leverage the adaptive intelligence of neural networks, we implement this operator through a parameterized network that enables the nonlinear synthesis of basis functions into the final solution.
\[\mathcal{K}_\theta: \{\underbrace{\{f_i(x)\}_{i=1}^m  \{G_j(t)\}_{j=1}^n}_{\substack{\text{Basis Function} \\ \text{Space}}}\} \xrightarrow{\text{Neural Synthesis}} U\] 
where $\theta$ represents the trainable network parameters, and $\mathcal{K}_\theta$ achieves the following. 1) Nonlinear basic coupling: Through activation functions, $\delta(\cdot),\mathcal{K}_\theta$ establishes cross-basis interactions, enabling nonlinear combinations beyond the linear superposition $\sum c_{ij}f_i(x)g_j(t)$. 2) Dynamically adjusted basis coefficients via hidden-layer states: $$ u(x,t)=\phi(W^L\delta(W^{L-1}\cdots \delta(w^1(f_i(x),g_j(t))+b^1)+b^{L-1})+b^L)$$, enabling spatiotemporal context-aware synthesis.
\item $f_1(x),\cdots, f_m(x)$: Spatially parameterized basis functions are formed under predefined spectral constraints. Example: Trigonometric bases for spatial dimensions:
\[f_i = a_i \sin(\omega_i x + \phi_i) + b_i\]

\item $g_1(t),\cdots,g_n(t)$: Temporally modulated basis functions are formed with adaptive localization. Example: Gaussian bases for temporal dynamics:
\[g_j(t) = \alpha_jexp(-\frac{(t-\mu_j)^2}{2\sigma^2_j})\]
\end{itemize}
$a_i, \omega_i, \phi_i, b_i,\alpha_j,\mu_j,\sigma_j\in \mathbb{R}$ are the learnable parameters of the network.

Compared with the conventional DNN methods, our network presents a more universal PDE-solving framework through explicitly designed basis functions. Its key advantages include the following.
\begin{enumerate}[1]
\item Enhanced generality via finite series approximation implemented over single-network closed-form solution fitting.
\item Customizable basis functions guided by intrinsic PDE properties, enabling effective analyses of the learned solution representations.
\item A per-point functional series fitting scheme that enriches the topological structural information of solutions, thereby granting the GEN superior robustness and extensibility.
\end{enumerate}


\section{Methods}\label{sec4}
\subsection{Basis function selection}

The methodological selection of trigonometric and Gaussian functions as the dual basis functions employed in this study is guided by the following considerations. First, Gaussian functions, whose original purpose was to use characteristic functions, can be expressed as a combination of characteristic functions. This method better explains the weak extension and local consistency of the traditional point-by-point DNN fitting approach. However, the discontinuity of the underlying characteristic functions leads to inappropriate derivative requirements due to the use of DNN method. Selecting a basis function that is similar to the target characteristic function is necessary for explaining and understanding our method, so the differentiability of Gaussian functions is used for the experiments. Other functions, such as quadratic functions, can also be selected. Second, trigonometric functions are selected because they possess a property similar to that of the Fourier series form, which provides the following advantages. 1) Domain universality: These functions are defined over $\mathbb{R}$ with infinite differentiability. 2) Orthogonality of basis functions: Trigonometric functions are orthogonal over a given period. 3) Completeness: Trigonometric function series can represent any piecewise-smooth periodic function (Dirichlet conditions). 4) PDE solution representation capacity: Solutions frequently involving trigonometric series expansions, aligning with the fundamental PDE theory.

Our methodological paradigm systematically embeds domain-specific physical priors by exploiting the intrinsic physical properties of differential equations by conducting structured basis function engineering within spatiotemporal frameworks. This form enables a physics-informed constraint integration process guided by the conservation principles and dynamic evolution characteristics that are inherent to each PDE class. {\bf Heat equation implementation:} Given a priori knowledge regarding exponential temporal decay characteristics, we restrict the basis construction procedure to spatial dimensions. This operational constraint motivates trigonometric or Gaussian basis functions designed to capture the diffusion process in space while inherently preserving the temporally asymptotic decay structure. {\bf Wave equation formulation: } The hyperbolic nature of the wave equation manifests through its characteristic propagation structure, which is governed by the d'Alembert solution framework. This intrinsic property motivates our construction of directional composite basis functions $\psi_{\pm}(\xi) = \phi_1(x - ct) + \phi_2(x + ct)$, where $\xi = x \pm ct$ denotes the characteristic coordinates. Such characteristic-aligned function compositions inherently preserve the fundamental duality of travelling wave solutions while maintaining strict adherence to the finite propagation speed constraint of the equation $|dx/dt| = c$. {\bf Burgers' equation exploration: } In the absence of strong physical priors, we adopt a comprehensive testing framework employing hybrid bases across both the spatial and temporal domains.

\subsection{Model training}
The general PDE form can be expressed as
\begin{equation}
\begin{aligned}
u_t+\mathcal{F}(x,u(x,t)) &=& f(x,t), x\in\Omega,t\in[0,T]\\
\mathcal{B}(u) &=& b(x,t), x\in\partial \Omega\\
\mathcal{I}(u)&=&i(x,t) , t= 0
\end{aligned}
\label{geneq}
\end{equation}
where $u(x,t)$ is the latent solution to be decided, $u_t$ is the temporal derivative, $\mathcal{F} (\cdot)$ is the linear or nonlinear spatial differential operator containing the possible orders of spatial derivatives, $f(x,t)$ is the source term, $\mathcal{B}(\cdot)$ is the boundary operator for calculating the boundary values, $b(x,t)$ is the boundary condition, $\mathcal{I}(\cdot)$ is the initial operator for calculating the initial values, $i(x,t)$ is the initial condition, $\Omega$ is the computational domain and $\partial \Omega$ is the boundary.

Considering the PDEs in the form of equation (\ref{geneq}) and inputting the spatial coordinate $x$ and the temporal coordinate $t$, we can obtain the corresponding values of a series of basis functions $f_i(x)$ and $g_j(t)$, which are input into the synthetic network to obtain the network prediction solution $\hat{u}$. By properly designing the loss function and a certain optimization algorithm, we can finally obtain a solution to which the network converges. Three points need to be explained here.
\begin{enumerate}
\item {\bf Basis function initialization protocols: } \\
Trigonometric function:
\begin{align}
    a_i &\sim \mathcal{U}(0,1) \\
    \omega_i &\sim i\pi \cdot \mathcal{U}(0,1)
\end{align}
Gaussian function
\begin{align}
    a_i &\sim \mathcal{U}(0,1) \\
    \sigma_i &\sim \mathcal{U}(0,1) \\
    \mu_i &\sim \min + (\max-\min) \cdot \mathcal{U}(0,1)
\end{align}
where $\mathcal{U}(0,1)$ is a uniform distribution and $\min$ and $\max$ are the minimum and maximum values of the current coordinates, respectively.

\item {\bf Network synthesis architecture specifications: }
\begin{itemize}
    \item Input dimensionality: $M+N$ (basis concatenation)
    \item Hidden layer: 20 neurons with $tanh$ activation
    \item Output: Linear transformation
\end{itemize}
\item {\bf Physics-informed loss function}: Aligned with conventional PINN frameworks:
\begin{equation}
\mathcal{L} = \underbrace{\mathbb{E}_{(x,t)}\left[\left(\mathcal{N}[\hat{u}]\right)^2\right]}_{\text{MSE}_{\text{PDE}}} + \lambda\underbrace{\mathbb{E}_{\partial\Omega}\left[(\hat{u}-u_{\text{BC}})^2\right]}_{\text{MSE}_{\text{BC}}} + \gamma\underbrace{\mathbb{E}_{t=0}\left[(\hat{u}-u_0)^2\right]}_{\text{MSE}_{\text{IC}}}
\end{equation}

\end{enumerate}

The GEN model is built using the PyTorch framework \cite{paszke2019pytorch,paszke2017automatic,ketkar2021automatic} and trained on one Nvidia RTX 4090 GPU for a total of 100000 iterations. The adaptive moment estimation (Adam) optimizer is used with the default parameters ($\beta_1=0.9$ and $\beta_2=0.999$). The learning rate is set to $1e-3$. 

\section{Results}\label{sec2}

We demonstrate our PDE-solving framework through three canonical model systems: the heat equation governing thermal diffusion, the wave equation describing vibrational propagation, and Burgers' equation for modelling nonlinear convection-diffusion phenomena. For different specific equations with various initial conditions and boundary conditions, we emphasize that prior information significantly influences the process of selecting appropriate basis functions, which directly impacts the accuracy and quality of the fitting results.

\subsection{Heat equation}

Simply, we commence our investigation with the heat equation, which is selected due to its well-characterized solution properties. The governing PDE is formulated as follows:
\begin{equation}
\begin{aligned}
u_t-u_{xx}=0\\
u(x,0)=sin(\frac{\pi}{2}x)\\
u(0,t)=u(2,t)=0
\end{aligned}
\end{equation}
The analytical solution admits the closed-form expression shown below:
\begin{equation}
u(x,t) = e^{-(\pi/2)^2 t} \sin\left(\frac{\pi}{2}x\right)
\end{equation}

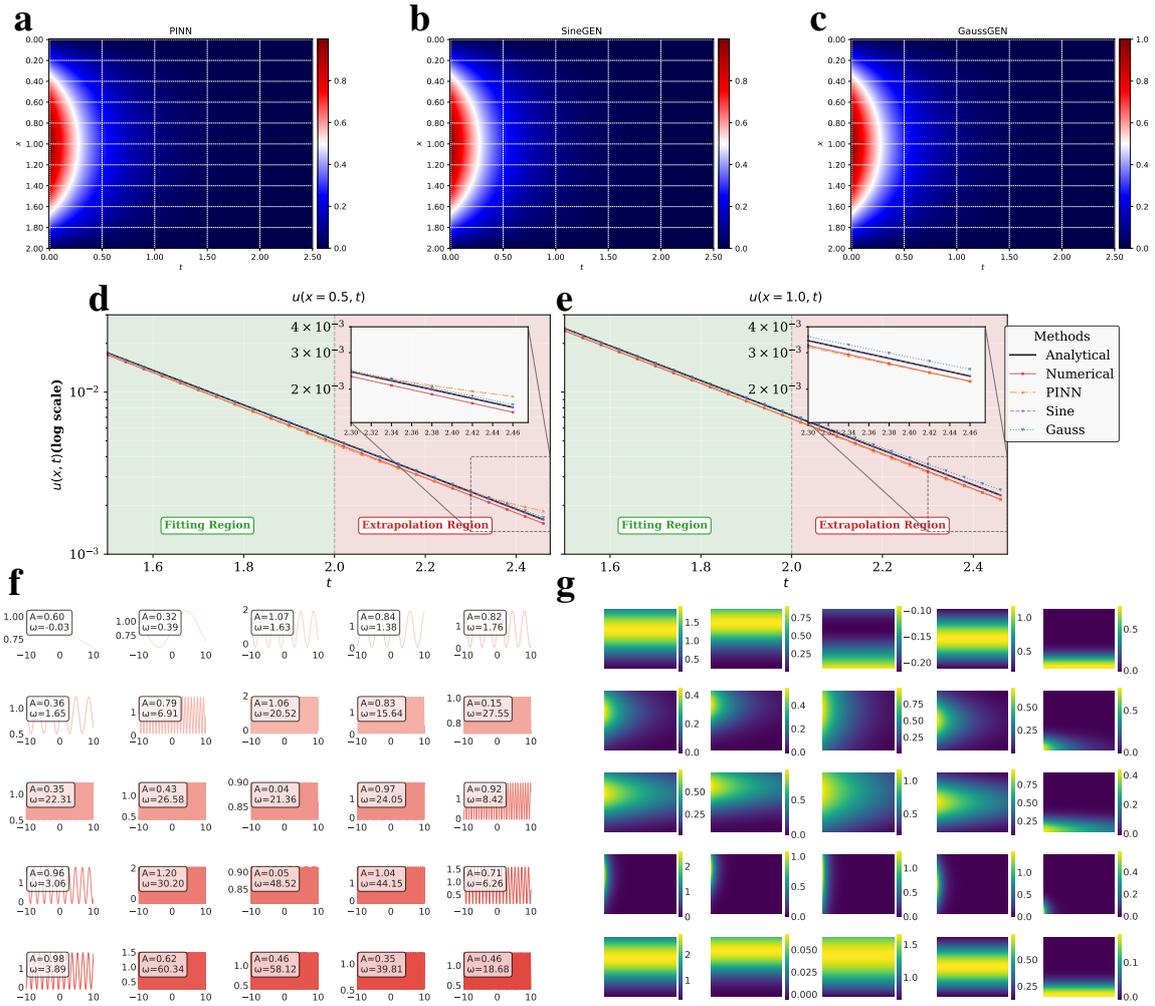
\begin{figure}
\centering
\includegraphics[width=6in]{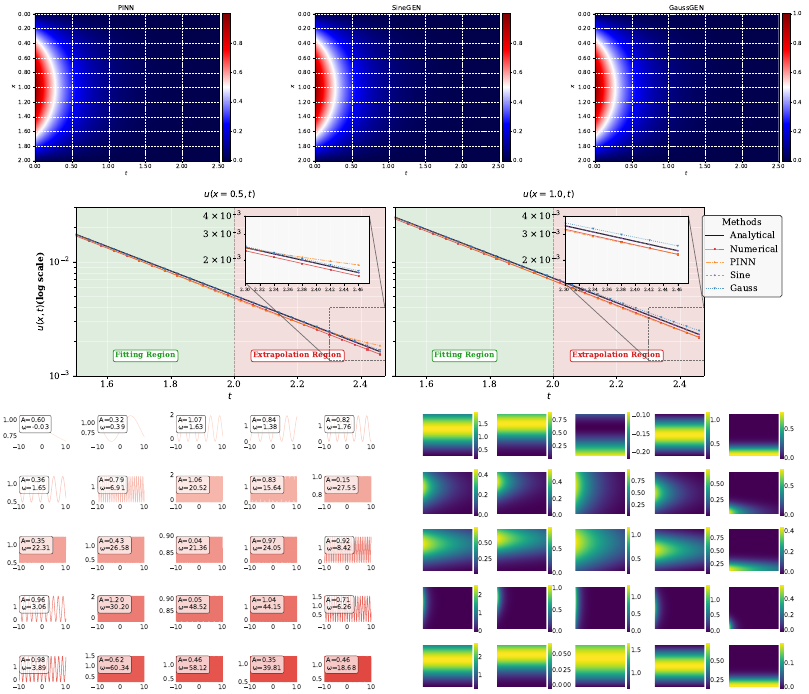}
\put(-428,368){\bf{\Large{a}}}
\put(-280,368){\bf{\Large{b}}}
\put(-130,368){\bf{\Large{c}}}
\put(-400,262){\bf{\Large{d}}}
\put(-225,262){\bf{\Large{e}}}
\put(-430,155){\bf{\Large{f}}}
\put(-225,155){\bf{\Large{g}}}
\caption{Comparison among solutions for the heat equation: (a) The PINN. (b-c) The SineGEN and GaussGEN developed under the proposed GEN framework. (d-e) The temporal profiles produced at x = 0.5 (left) and x = 1.0 (right), demonstrating increasing extrapolation errors beyond the training domain ($t > 2.0$), which is particularly evident in the PINN predictions. (f-g) The 25 basis functions learned by the two GEN methods.}
\label{heat_res}
\end{figure}

Two novel basis function schemes are developed for this study:
\begin{itemize}
\item \textbf{SineGEN}: Trigonometric basis functions for spatial positions with $N=25$:
$$\{a_ie^{-\omega_i^2t}sin(\omega_ix)+b\}_{i=1}^N$$
\item \textbf{GaussGEN}: Gaussian basis functions for spatial positions with $m=n=5$:
$$\{e^{-\mu_{tj}t}\}_{j=1}^n \otimes \{a_ie^{(\frac{x-\mu_{xi}}{\sigma_i})^2}\}_{i=1}^{m}$$
\end{itemize}
A comparative analysis between the PINN and the conventional numerical solvers is presented in Fig. \ref{heat_res}, which demonstrates the efficacy of our methodology.

From the solution results, visually, both the PINN and the GENs with two different basis functions achieve satisfactory outcomes. A further analysis performed through appropriate extrapolations from the ixed positions reveals that although our equation possesses an explicit closed-form solution, the PINN exhibits significant deviations outside the original domain. This indicates that the pursuit of a "black-box" explicit closed-form solution by the PINN fails to attain the expected true solution and instead produces only a local fitting result. In contrast, when our method formally aligns with the solution structure, it can ultimately yield expressions with better extrapolation capabilities and higher accuracy or approximate the true solution. Even with poorly chosen basis functions, our method achieves an accuracy comparable to that of the PINN, although its extrapolation performance depends on the properties of the chosen basis functions. Figs. \ref{heat_res}(f-g) illustrate specific characteristics of the basis functions, which not only aid in further analyses but also increase the resulting solution accuracy. For example, analysing frequencies and amplitudes through techniques that are analogous to Fourier transformations could further refine these insights.

\subsection{Wave equation}
\begin{figure}
\centering
\includegraphics[width=6in]{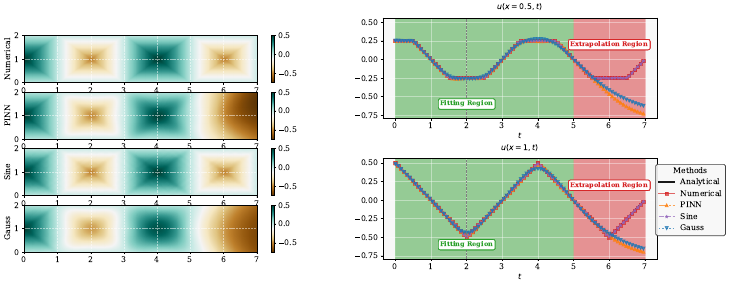}
\put(-430,155){\bf{\Large{a}}}
\put(-225,163){\bf{\Large{b}}}
\put(-225,80){\bf{\Large{c}}}
\caption{Wave equation: (a) Heatmap comparison among the numerical solution, the PINN and the two GEN results, with the colour intensities representing the solution values. (b) and (c) Temporal snapshots of the solutions produced at spatial locations $x=0.5$ and $x=3.0$, respectively. The green shaded area represents the fitting region (training domain), and the red shaded area represents the extrapolation region.}
\label{wave_res}
\end{figure}

Next, we further demonstrate the application of the GEN to a wave equation. Let us consider the following wave equation:
\begin{equation}
\begin{aligned}
u_{tt} - u_{xx} = 0 \\
u(x,0)=
    \begin{cases}
        \frac{x}{2}, &x<1 \\
        1-\frac{x}{2}, &x\geq1
    \end{cases}\\
u(0,t)=u(2,t) = 0
\end{aligned}
\end{equation}
For this equation, two characteristic lines exist: $x-t=c_1$ and $x+t=c_2$. On the basis of this property, we design basis functions with the forms of $\zeta(x+t)$ and $\eta (x-t)$. Similarly, we also test trigonometric functions $\{a_i\sin(\omega(x-t))+b_i\sin(\omega(x+t))+d\}_{i=1}^{25}$ and Gaussian basis functions $\{a_i\exp((\frac{x-\mu_{xi}-t-\mu_{ti}}{\sigma_{xi}\sigma_{ti}})^2)+b_i \exp((\frac{x+\mu_{xi}-t-\mu_{ti}}{\sigma_{xi}\sigma_{ti}})^2)+d\}_{i=1}^{25}$ for solving the equation. The results are illustrated in Fig. \ref{wave_res}.

As above, the experimental results show that within the training domain, both the PINN and the GEN methods with trigonometric basis functions produce satisfactory numerical solutions. However, outside the fitting region, owing to the nature of the periodic extension, the PINN and Gaussian basis functions exhibit ill-posed behaviour, whereas the inherent periodicity of the trigonometric functions effectively preserves the periodic extension. Additionally, at the fixed spatial point $x=1$, the Gaussian basis functions retain the smooth transition at the peak rather than a sharp turning point. On the one hand, this highlights that the appropriate selection of basis functions can enhance the accuracy of the solutions output by a network. On the other hand, it underscores the critical role of human prior knowledge in PDE-solving tasks. These insights are vital for conducting robust performance analyses when applying DNN methods to solve PDEs in real-world scenarios.

\subsection{Burgers' equation}
\begin{figure}
\centering
\includegraphics[width=6in]{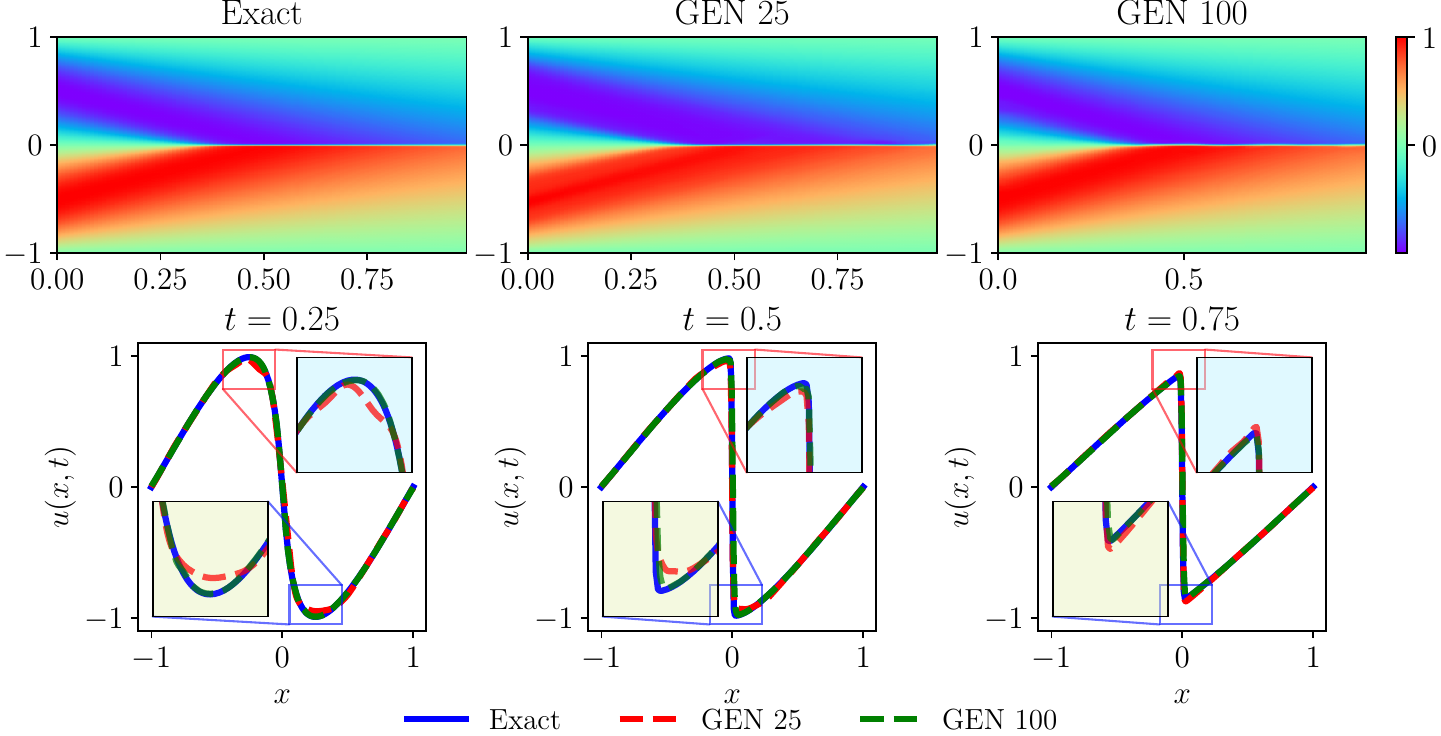}
\put(-425,220){\bf{\Large{a}}}
\put(-287,220){\bf{\Large{b}}}
\put(-143,220){\bf{\Large{c}}}
\put(-425,120){\bf{\Large{d}}}
\put(-287,120){\bf{\Large{e}}}
\put(-143,120){\bf{\Large{f}}}
\caption{Burgers' equation: (a)-(c) Heatmap comparisons between exact solutions and GENs with various numbers of basis functions. (d)-(f) Spatial snapshots of the solutions produced at temporal positions $t=0.25, 0.5$ and $0.75$, respectively. Magnified views near the maxima and minima are shown in the upper-right and lower-left corners, respectively. Marked fitting accuracy discrepancies are observed when a smaller number of basis functions is employed.}
\label{burgers}
\end{figure}
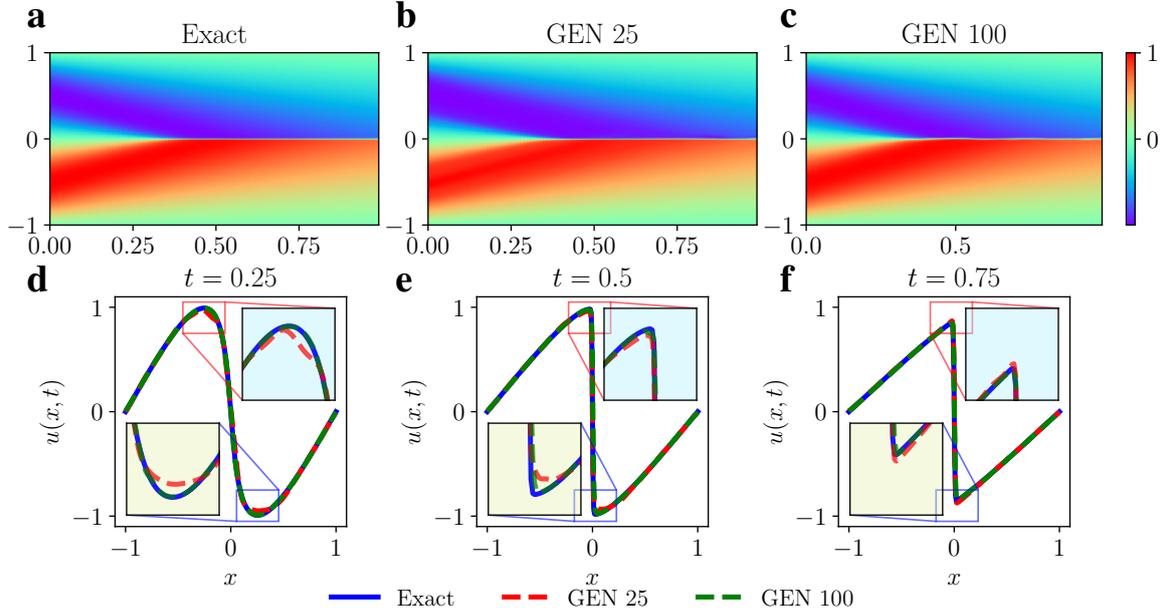

In previous discussions, we observed that the method of approximating PDEs using trigonometric functions appears to outperform PINNs. This advantage likely stems from the fact that any periodic function satisfying the Dirichlet conditions can be represented as a superposition of sine (or cosine) functions with varying frequencies. To explore this concept further, we conduct a study using trigonometric basis functions for solving Burgers' equation, which are defined as follows:
\begin{equation}
\begin{aligned}
u_t+uu_x-\frac{0.01}{\pi}u_{xx} = 0\\
u(x,0) = -\sin(\pi x)\\
u(1,t)=u(-1,t) = 0
\end{aligned}
\end{equation}
Here, we conduct experiments using the same sine basis functions as those applied for the wave equation but with varying numbers of basis functions. Fig. \ref{burgers} presents the solution results obtained with 25 basis functions (GEN 25) and 100 basis functions (GEN 100).

A summary of our results obtained for this example is presented in Fig. \ref{burgers}. A particularly intriguing finding is that our sine-based basis functions achieve high accuracy in terms of solving the equation regardless of the number of basis functions used. However, as observed in the localized magnified plots near the maxima and minima of the temporal snapshots, a certain degree of error persists when fewer basis functions are employed (e.g., GEN 25), whereas higher precision is attained with an increased number of basis functions (e.g., GEN 100). This phenomenon suggests that our method achieves large-scale fitting (a globally optimal solution) with fewer basis functions, while incorporating more basis functions enhances the resolution of the localized details, leading to superior fine-grained feature capturing accuracy.

\section{Discussion}\label{sec3}

The traditional DNNs for solving PDEs face critical limitations. 1) Robustness deficits: Data-driven methods are noise-sensitive approaches, with their extrapolation errors growing exponentially. 2) Limited extensibility: Black-box models struggle to incorporate prior knowledge (e.g., symmetries and conservation laws) and fail to achieve continuity. A novel paradigm for solving PDEs through the explicit construction of solutions via equation-specific customized basis functions is introduced in this paper, ensuring enhanced fidelity to the intrinsic properties and governing laws of the true solutions. This explicit synthesis framework offers three key advantages.
\begin{itemize}
\item {\bf Series representations enable regional extensions:} A finite-term series expansion method based on basis functions uses the definition domain and global analytic relationships of these functions to achieve local continuity, which ensures stability. Moreover, basis functions inherently support domain extensions, whereas appropriately selected basis functions can enable DNN-based methods to solve PDE problems and align more closely with the global solution rather than being confined solely to the training interval.

\item {\bf PED-driven basis function design:} The conventional DNN-based methods for solving most equations exhibit inherent limitations, primarily manifesting in scenarios where critical physical properties, such as symmetry, periodicity, or specific intrinsic features, are not explicitly embedded into the network architectures. For example, in wave equations, while periodicity is a defining characteristic, the traditional point-to-point mapping paradigms often fail to reliably generalize to nontraining regions. Addressing this defect, the incorporation of periodic basis functions into the proposed network architecture proves effective for mitigating such shortcomings.

\item {\bf Explicit solution structure analysis:} Analytical representations of basis functions permit systematic characterization and targeted modulation processes to be implemented, with trigonometric spectral analysis enabling dimensionality reduction to be achieved through the execution of pruning operations within optimization frameworks.

\end{itemize}

While the proposed method demonstrates significant advancements in terms of solving PDEs, three critical limitations hinder its broader applicability and performance optimization potential.

\begin{itemize}
\item {\bf Basis function selection:} The selection of appropriate basis functions that are tailored to specific differential equations fundamentally determines the solution accuracy our method. While our selection of trigonometric and Gaussian functions is guided by empirical heuristics (as discussed in the methodology section), the results section critically demonstrates that customized basis functions conforming to particular physical principles and system properties are essential for achieving optimal performance.
\item {\bf Slow convergence during training:} Experimental observations reveal a protracted training convergence process, with our implementation requiring 100,000 iterations, which is significantly greater than that of conventional PINNs. This computational intensiveness presents a major challenge for achieving efficient and explicit representations, necessitating the development of accelerated optimization algorithms.
\item{\bf Trade-offs regarding the number of basis functions:} The solution precision our approach is sensitive to the number of selected basis functions. Insufficient cardinality degrades the accuracy of the network, whereas the use of excessive functions induces parameter redundancy. The inability of the current framework to autonomously adapt its basis function quantity through an intelligent cardinality-parameter co-optimization scheme constitutes a notable limitation with respect to its use in high-precision applications.
\end{itemize}

Of course, the primary limitation of this paper is that I am not engaged in research related to PDEs. The selection of basis functions mentioned in this work may not necessarily be reasonable or optimal, and it requires subsequent PDE researchers to continually explore and identify suitable basis functions, or to integrate them into neural networks to enhance applications across more fields. The idea behind this paper originated three years ago, and I have decided not to invest further time in refining it. I hope that someone with the right interest will uncover and further develop the techniques presented here, continuously improving the effectiveness of the method.

In conclusion, a novel paradigm for constructing explicit PDE solutions is proposed in this study. The developed methodology demonstrates enhanced model extensibility and robustness while establishing an analytical framework for DNN-based PDE solvers. This advancement is achieved through the systematic integration of prior knowledge via basis functions as architectural constraints during the solution formulation process. A comprehensive experimental validation confirms the ability of the proposed approach to resolve PDEs with high precision.

\section{Acknowledgments}
This research is supported by National Natural Science Foundation of China, Grant/Award Numbers (12426308) and the National Key Research and Development Program of China, Grant/Award Number (2023YFA1011402), Beijing Postdoctoral Research Foundation and the Key Research Project of the Academy for Multidisciplinary Studies, Capital Normal University. The authors are also grateful to National Center for Applied Mathematics Beijing for funding this research work.

\bibliographystyle{elsarticle-harv} 
\bibliography{ref}

@article{raissi2019physics,
  title={Physics-informed neural networks: A deep learning framework for solving forward and inverse problems involving nonlinear partial differential equations},
  author={Raissi, Maziar and Perdikaris, Paris and Karniadakis, George E},
  journal={Journal of Computational physics},
  volume={378},
  pages={686--707},
  year={2019},
  publisher={Elsevier}
}

@article{wang2024pinn,
  title={NAS-PINN: Neural architecture search-guided physics-informed neural network for solving PDEs},
  author={Wang, Yifan and Zhong, Linlin},
  journal={Journal of Computational Physics},
  volume={496},
  pages={112603},
  year={2024},
  publisher={Elsevier}
}

@article{thuerey2021physics,
  title={Physics-based deep learning},
  author={Thuerey, Nils and Holl, Philipp and Mueller, Maximilian and Schnell, Patrick and Trost, Felix and Um, Kiwon},
  journal={arXiv preprint arXiv:2109.05237},
  year={2021}
}

@article{brunton2024promising,
  title={Promising directions of machine learning for partial differential equations},
  author={Brunton, Steven L and Kutz, J Nathan},
  journal={Nature Computational Science},
  volume={4},
  number={7},
  pages={483--494},
  year={2024},
  publisher={Nature Publishing Group US New York}
}

@article{vinuesa2022enhancing,
  title={Enhancing computational fluid dynamics with machine learning},
  author={Vinuesa, Ricardo and Brunton, Steven L},
  journal={Nature Computational Science},
  volume={2},
  number={6},
  pages={358--366},
  year={2022},
  publisher={Nature Publishing Group US New York}
}

@article{karniadakis2021physics,
  title={Physics-informed machine learning},
  author={Karniadakis, George Em and Kevrekidis, Ioannis G and Lu, Lu and Perdikaris, Paris and Wang, Sifan and Yang, Liu},
  journal={Nature Reviews Physics},
  volume={3},
  number={6},
  pages={422--440},
  year={2021},
  publisher={Nature Publishing Group UK London}
}

@article{cuomo2022scientific,
  title={Scientific machine learning through physics--informed neural networks: Where we are and what’s next},
  author={Cuomo, Salvatore and Di Cola, Vincenzo Schiano and Giampaolo, Fabio and Rozza, Gianluigi and Raissi, Maziar and Piccialli, Francesco},
  journal={Journal of Scientific Computing},
  volume={92},
  number={3},
  pages={88},
  year={2022},
  publisher={Springer}
}

@article{hornik1989multilayer,
  title={Multilayer feedforward networks are universal approximators},
  author={Hornik, Kurt and Stinchcombe, Maxwell and White, Halbert},
  journal={Neural networks},
  volume={2},
  number={5},
  pages={359--366},
  year={1989},
  publisher={Elsevier}
}

@article{kovachki2023neural,
  title={Neural operator: Learning maps between function spaces with applications to pdes},
  author={Kovachki, Nikola and Li, Zongyi and Liu, Burigede and Azizzadenesheli, Kamyar and Bhattacharya, Kaushik and Stuart, Andrew and Anandkumar, Anima},
  journal={Journal of Machine Learning Research},
  volume={24},
  number={89},
  pages={1--97},
  year={2023}
}

@inproceedings{anandkumar2020neural,
  title={Neural operator: Graph kernel network for partial differential equations},
  author={Anandkumar, Anima and Azizzadenesheli, Kamyar and Bhattacharya, Kaushik and Kovachki, Nikola and Li, Zongyi and Liu, Burigede and Stuart, Andrew},
  booktitle={ICLR 2020 workshop on integration of deep neural models and differential equations},
  year={2020}
}

@article{rosofsky2023applications,
  title={Applications of physics informed neural operators},
  author={Rosofsky, Shawn G and Al Majed, Hani and Huerta, EA},
  journal={Machine Learning: Science and Technology},
  volume={4},
  number={2},
  pages={025022},
  year={2023},
  publisher={IOP Publishing}
}

@article{lu2021learning,
  title={Learning nonlinear operators via DeepONet based on the universal approximation theorem of operators},
  author={Lu, Lu and Jin, Pengzhan and Pang, Guofei and Zhang, Zhongqiang and Karniadakis, George Em},
  journal={Nature machine intelligence},
  volume={3},
  number={3},
  pages={218--229},
  year={2021},
  publisher={Nature Publishing Group UK London}
}

@article{li2020multipole,
  title={Multipole graph neural operator for parametric partial differential equations},
  author={Li, Zongyi and Kovachki, Nikola and Azizzadenesheli, Kamyar and Liu, Burigede and Stuart, Andrew and Bhattacharya, Kaushik and Anandkumar, Anima},
  journal={Advances in Neural Information Processing Systems},
  volume={33},
  pages={6755--6766},
  year={2020}
}

@inproceedings{lifourier,
  title={Fourier Neural Operator for Parametric Partial Differential Equations},
  author={Li, Zongyi and Kovachki, Nikola Borislavov and Azizzadenesheli, Kamyar and Bhattacharya, Kaushik and Stuart, Andrew and Anandkumar, Anima and others},
  booktitle={International Conference on Learning Representations}
}

@article{cao2024laplace,
  title={Laplace neural operator for solving differential equations},
  author={Cao, Qianying and Goswami, Somdatta and Karniadakis, George Em},
  journal={Nature Machine Intelligence},
  volume={6},
  number={6},
  pages={631--640},
  year={2024},
  publisher={Nature Publishing Group UK London}
}

@article{brandstetter2025envisioning,
  title={Envisioning better benchmarks for machine learning PDE solvers},
  author={Brandstetter, Johannes},
  journal={Nature Machine Intelligence},
  volume={7},
  number={1},
  pages={2--3},
  year={2025},
  publisher={Nature Publishing Group}
}

@article{mcgreivy2024weak,
  title={Weak baselines and reporting biases lead to overoptimism in machine learning for fluid-related partial differential equations},
  author={McGreivy, Nick and Hakim, Ammar},
  journal={Nature Machine Intelligence},
  volume={6},
  number={10},
  pages={1256--1269},
  year={2024},
  publisher={Nature Publishing Group UK London}
}

@article{M2025Machine,
  title={Machine learning solutions looking for PDE problems},
  author={},
  journal={Nature Machine Intelligence},
  volume={7},
  number={1},
  pages={1},
  year={2025},
  publisher={Nature Publishing Group UK London}
}

@article{chuang2022experience,
  title={Experience report of physics-informed neural networks in fluid simulations: pitfalls and frustration},
  author={Chuang, Pi-Yueh and Barba, Lorena A},
  journal={arXiv preprint arXiv:2205.14249},
  year={2022}
}

@article{chuang2023predictive,
  title={Predictive limitations of physics-informed neural networks in vortex shedding},
  author={Chuang, Pi-Yueh and Barba, Lorena A},
  journal={arXiv preprint arXiv:2306.00230},
  year={2023}
}

@article{wang2022and,
  title={When and why PINNs fail to train: A neural tangent kernel perspective},
  author={Wang, Sifan and Yu, Xinling and Perdikaris, Paris},
  journal={Journal of Computational Physics},
  volume={449},
  pages={110768},
  year={2022},
  publisher={Elsevier}
}

@article{krishnapriyan2021characterizing,
  title={Characterizing possible failure modes in physics-informed neural networks},
  author={Krishnapriyan, Aditi and Gholami, Amir and Zhe, Shandian and Kirby, Robert and Mahoney, Michael W},
  journal={Advances in neural information processing systems},
  volume={34},
  pages={26548--26560},
  year={2021}
}

@inproceedings{basir2022critical,
  title={Critical investigation of failure modes in physics-informed neural networks},
  author={Basir, Shamsulhaq and Senocak, Inanc},
  booktitle={AiAA SCITECH 2022 Forum},
  pages={2353},
  year={2022}
}

@article{karnakov2024solving,
  title={Solving inverse problems in physics by optimizing a discrete loss: Fast and accurate learning without neural networks},
  author={Karnakov, Petr and Litvinov, Sergey and Koumoutsakos, Petros},
  journal={PNAS nexus},
  volume={3},
  number={1},
  pages={pgae005},
  year={2024},
  publisher={Oxford University Press US}
}

@article{chen2025pf,
  title={PF-PINNs: Physics-informed neural networks for solving coupled Allen-Cahn and Cahn-Hilliard phase field equations},
  author={Chen, Nanxi and Lucarini, Sergio and Ma, Rujin and Chen, Airong and Cui, Chuanjie},
  journal={Journal of Computational Physics},
  pages={113843},
  year={2025},
  publisher={Elsevier}
}

@article{yu2024sinc,
  title={Sinc kolmogorov-arnold network and its applications on physics-informed neural networks},
  author={Yu, Tianchi and Qiu, Jingwei and Yang, Jiang and Oseledets, Ivan},
  journal={arXiv preprint arXiv:2410.04096},
  year={2024}
}

@article{zhang2022fourier,
  title={Fourier neural operator for solving subsurface oil/water two-phase flow partial differential equation},
  author={Zhang, Kai and Zuo, Yuande and Zhao, Hanjun and Ma, Xiaopeng and Gu, Jianwei and Wang, Jian and Yang, Yongfei and Yao, Chuanjin and Yao, Jun},
  journal={Spe Journal},
  volume={27},
  number={03},
  pages={1815--1830},
  year={2022},
  publisher={OnePetro}
}

@article{li2020fourier,
  title={Fourier neural operator for parametric partial differential equations},
  author={Li, Zongyi and Kovachki, Nikola and Azizzadenesheli, Kamyar and Liu, Burigede and Bhattacharya, Kaushik and Stuart, Andrew and Anandkumar, Anima},
  journal={arXiv preprint arXiv:2010.08895},
  year={2020}
}

@article{qi2024gabor,
  title={Gabor-filtered fourier neural operator for solving partial differential equations},
  author={Qi, Kai and Sun, Jian},
  journal={Computers \& Fluids},
  volume={274},
  pages={106239},
  year={2024},
  publisher={Elsevier}
}

@article{wang2021eigenvector,
  title={On the eigenvector bias of Fourier feature networks: From regression to solving multi-scale PDEs with physics-informed neural networks},
  author={Wang, Sifan and Wang, Hanwen and Perdikaris, Paris},
  journal={Computer Methods in Applied Mechanics and Engineering},
  volume={384},
  pages={113938},
  year={2021},
  publisher={Elsevier}
}

@article{jin2024fourier,
  title={Fourier warm start for physics-informed neural networks},
  author={Jin, Ge and Wong, Jian Cheng and Gupta, Abhishek and Li, Shipeng and Ong, Yew-Soon},
  journal={Engineering Applications of Artificial Intelligence},
  volume={132},
  pages={107887},
  year={2024},
  publisher={Elsevier}
}

@article{bounnah2025physics,
  title={Physics informed neural network with Fourier feature for natural convection problems},
  author={Bounnah, Younes and Mihoubi, Mustapha Kamel and Larbi, Salah},
  journal={Engineering Applications of Artificial Intelligence},
  volume={146},
  pages={110327},
  year={2025},
  publisher={Elsevier}
}

@article{song2023simulating,
  title={Simulating seismic multifrequency wavefields with the Fourier feature physics-informed neural network},
  author={Song, Chao and Wang, Yanghua},
  journal={Geophysical Journal International},
  volume={232},
  number={3},
  pages={1503--1514},
  year={2023},
  publisher={Oxford University Press}
}

@article{cai2021physics,
  title={Physics-informed neural networks (PINNs) for fluid mechanics: A review},
  author={Cai, Shengze and Mao, Zhiping and Wang, Zhicheng and Yin, Minglang and Karniadakis, George Em},
  journal={Acta Mechanica Sinica},
  volume={37},
  number={12},
  pages={1727--1738},
  year={2021},
  publisher={Springer}
}

@article{sharma2023review,
  title={A review of physics-informed machine learning in fluid mechanics},
  author={Sharma, Pushan and Chung, Wai Tong and Akoush, Bassem and Ihme, Matthias},
  journal={Energies},
  volume={16},
  number={5},
  pages={2343},
  year={2023},
  publisher={MDPI}
}

@article{wu2023comprehensive,
  title={A comprehensive study of non-adaptive and residual-based adaptive sampling for physics-informed neural networks},
  author={Wu, Chenxi and Zhu, Min and Tan, Qinyang and Kartha, Yadhu and Lu, Lu},
  journal={Computer Methods in Applied Mechanics and Engineering},
  volume={403},
  pages={115671},
  year={2023},
  publisher={Elsevier}
}

@article{wei2022small,
  title={Small-data-driven fast seismic simulations for complex media using physics-informed Fourier neural operators},
  author={Wei, Wei and Fu, Li-Yun},
  journal={Geophysics},
  volume={87},
  number={6},
  pages={T435--T446},
  year={2022},
  publisher={Society of Exploration Geophysicists}
}

@article{sallam2023use,
  title={On the use of Fourier Features-Physics Informed Neural Networks (FF-PINN) for forward and inverse fluid mechanics problems},
  author={Sallam, Omar and F{\"u}rth, Mirjam},
  journal={Proceedings of the Institution of Mechanical Engineers, Part M: Journal of Engineering for the Maritime Environment},
  volume={237},
  number={4},
  pages={846--866},
  year={2023},
  publisher={SAGE Publications Sage UK: London, England}
}

@article{li2023phase,
  title={Phase-Field DeepONet: Physics-informed deep operator neural network for fast simulations of pattern formation governed by gradient flows of free-energy functionals},
  author={Li, Wei and Bazant, Martin Z and Zhu, Juner},
  journal={Computer Methods in Applied Mechanics and Engineering},
  volume={416},
  pages={116299},
  year={2023},
  publisher={Elsevier}
}

@article{li2024tutorials,
  title={Tutorials: Physics-informed machine learning methods of computing 1D phase-field models},
  author={Li, Wei and Fang, Ruqing and Jiao, Junning and Vassilakis, Georgios N and Zhu, Juner},
  journal={APL Machine Learning},
  volume={2},
  number={3},
  year={2024},
  publisher={AIP Publishing}
}

@article{jacot2018neural,
  title={Neural tangent kernel: Convergence and generalization in neural networks},
  author={Jacot, Arthur and Gabriel, Franck and Hongler, Cl{\'e}ment},
  journal={Advances in neural information processing systems},
  volume={31},
  year={2018}
}

@inproceedings{huang2021fl,
  title={Fl-ntk: A neural tangent kernel-based framework for federated learning analysis},
  author={Huang, Baihe and Li, Xiaoxiao and Song, Zhao and Yang, Xin},
  booktitle={International Conference on Machine Learning},
  pages={4423--4434},
  year={2021},
  organization={PMLR}
}

@article{liu2024kan,
  title={Kan: Kolmogorov-arnold networks},
  author={Liu, Ziming and Wang, Yixuan and Vaidya, Sachin and Ruehle, Fabian and Halverson, James and Solja{\v{c}}i{\'c}, Marin and Hou, Thomas Y and Tegmark, Max},
  journal={arXiv preprint arXiv:2404.19756},
  year={2024}
}

@article{wight2020solving,
  title={Solving Allen-Cahn and Cahn-Hilliard equations using the adaptive physics informed neural networks},
  author={Wight, Colby L and Zhao, Jia},
  journal={arXiv preprint arXiv:2007.04542},
  year={2020}
}

@article{paszke2019pytorch,
  title={Pytorch: An imperative style, high-performance deep learning library},
  author={Paszke, A},
  journal={arXiv preprint arXiv:1912.01703},
  year={2019}
}

@article{paszke2017automatic,
  title={Automatic differentiation in pytorch},
  author={Paszke, Adam and Gross, Sam and Chintala, Soumith and Chanan, Gregory and Yang, Edward and DeVito, Zachary and Lin, Zeming and Desmaison, Alban and Antiga, Luca and Lerer, Adam},
  year={2017}
}

@article{ketkar2021automatic,
  title={Automatic differentiation in deep learning},
  author={Ketkar, Nikhil and Moolayil, Jojo and Ketkar, Nikhil and Moolayil, Jojo},
  journal={Deep Learning with python: learn best practices of deep learning models with PyTorch},
  pages={133--145},
  year={2021},
  publisher={Springer}
}

\end{document}